\documentclass[eat,twocolumn]{jmlr}

\usepackage{longtable}

\usepackage{booktabs}
\usepackage[load-configurations=version-1]{siunitx} 
 
\usepackage{adjustbox}
\usepackage{multirow}
\usepackage{verbatim}
\usepackage{paralist}
\usepackage[compact]{titlesec}
\titlespacing{\section}{5pt}{2ex}{1ex}
\titlespacing{\subsection}{3pt}{1ex}{0ex}

\newcommand{\ecedit}[1]{\textcolor{black}{#1}}
\newcommand{\camr}[1]{\textcolor{black}{#1}}
\theorembodyfont{\upshape}
\theoremheaderfont{\scshape}
\theorempostheader{:}
\theoremsep{\newline}

\jmlrvolume{}
\firstpageno{1}

\jmlryear{2022}
\jmlrworkshop{Machine Learning for Health (ML4H) 2022}

\title[UniFL]{Universal EHR Federated Learning Framework}

\author{
\Name{Junu Kim} \Email{kjune0322@kaist.ac.kr}\\
\Name{Kyunghoon Hur} \Email{pacesun@kaist.ac.kr}\\
\Name{Seongjun Yang} \Email{seongjunyang@kaist.ac.kr}\\
\Name{Edward Choi} \Email{edwardchoi@kaist.ac.kr}\\
\addr KAIST
}

\begin{document}

 \maketitle

\begin{abstract}
Federated learning (FL) is the most practical multi-source learning method for electronic healthcare records (EHR).
Despite its guarantee of privacy protection, the wide application of FL is restricted by two large challenges: the heterogeneous EHR systems, and the non-i.i.d. data characteristic.
A recent research proposed a framework that unifies heterogeneous EHRs, named UniHPF.
We attempt to address both the challenges simultaneously by combining UniHPF and FL.
Our study is the first approach to unify heterogeneous EHRs into a single FL framework.
This combination provides an average of 3.4\% performance gain compared to local learning.
We believe that our framework is practically applicable in the real-world FL.

\end{abstract}


\section{Introduction}
\label{sec:intro}

\begin{figure*}[ht]
\floatconts
  {fig:prob_def}
  {\caption{\camr{The application of FL with EHR data is restricted by the two problems: (1) EHR system heterogeneity, and (2) non-i.i.d. problem. Although multiple prior researches attempt to resolve (2), there is no known solution for (1). Our framework is the first attempt to handle the both problems simultaneously by combining UniHPF and FL.}}}
  {\makebox[\textwidth][c]{\includegraphics[trim={120 250 110 180},clip,width=1.2\linewidth]{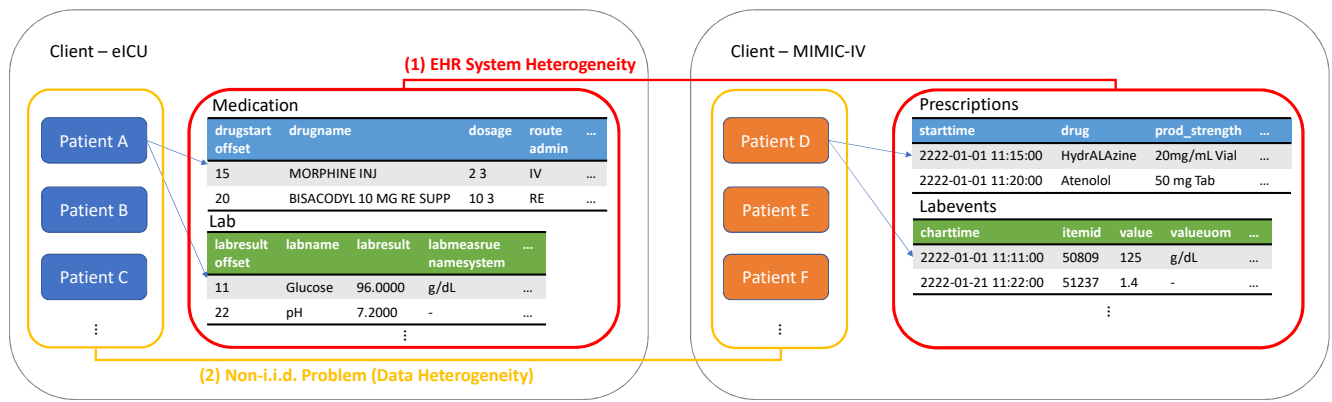}}}
\end{figure*}

Electronic healthcare record (EHR) is a rich data source that records whole hospital events for each patient.
Using EHR in machine learning (ML) allows \ecedit{us} to make useful predictions for patients' future health status \citep{doctorai,retain}.
Since the predictions can affect the life \camr{of the patients}, \ecedit{accurate prediction is vital.}
Due to the nature of ML, using more data is helpful to improve the model accuracy \citep{sordo2005sample, prusa2015effect}.
Considering that each EHR is maintained by each hospital, the size of the single-source data is limited.
Therefore, employing multiple hospitals' data (multi-source learning) is required to improve the accuracy.

Unfortunately, this is not \ecedit{straightforward} due to the privacy issue of EHR.
Since healthcare data contains personal information, exporting data outside the hospital is highly restricted.
Therefore, traditional centralized learning (CL) has limited practicality, because it need to gather all data into a central server (\algorithmref{alg:cl}).

In this situation, federated learning (FL) can be a solution since it does not need to share data among clients (hospital).
It only has to share model weights that are trained \ecedit{only} on each client data.
The global server aggregates the \ecedit{weights} and sends the \ecedit{global} model to each client \ecedit{in} each communication round \camr{(\algorithmref{alg:fed}).}
For this mechanism, FL is the most appropriate method for achieving multi-source learning on EHR while protecting patient privacy.

Despite \ecedit{its} benefits, the application \ecedit{of FL} is limited due to the heterogeneity of EHR system.
In \ecedit{each} EHR system (\textit{i.e.} client, hospital), the medical codes and the \ecedit{database} schema are typically not shared \camr{(\figureref{fig:prob_def} (1)).}
Therefore, most of the previous \ecedit{studies conduct FL experiments only with clients using} a single system \citep{lee2020federated, huang2019patient, fedpxn}.
However, these approaches are not able to handle the \camr{system heterogeneity of the real-world EHRs.}
Unifying all EHR systems into a standard format (common data model, CDM) can resolve this limitation \citep{rajkomar2018scalable, li2019distributed}.
However, it is not yet examined due to the cost- and time-consuming nature.

On the other hand, UniHPF \citep{unihpf} is a framework that can effectively handle heterogeneous EHR systems in the cost- and time-efficient manner.
It replaces medical codes with text and linearizes different database schemas to mutually compatible free text format (\figureref{fig:prob_def}).
However, \camr{the success of} UniHPF was only shown in the CL setting, which has the aforementioned practical limitations.
In FL, as opposed to CL, we have to consider the differences among the clients\ecedit{'} data distributions (non-i.i.d. problem, \figureref{fig:prob_def} (2)) \citep{rieke2020future, li2022federated}.

Therefore, we combine UniHPF with multiple FL methods to resolve the non-i.i.d. problem and compare among the methods.
Since the performance is increased in FL \camr{compared to without multi-source learning,} we successfully resolve both the privacy problem and the non-i.i.d. problem.
Our main contributions can be summarized as follows:

\begin{itemize}
    \item We suggest a practically applicable EHR multi-source learning framework by combining UniHPF and FL.
    \item Our proposed framework demonstrated improved prediction performance compared to local learning, and even occasionally showed similar performance to centralized learning.
    \item To the best of our knowledge, it is the first attempt to unify heterogeneous time-series EHRs into a single FL framework.
\end{itemize}

\begin{figure*}[!ht]
\floatconts
  {fig:unihpf}
  {\caption{\camr{Visualization of UniHPF \citep{unihpf}.
  UniHPF first makes the text representation of each event by linearizing the schema and replacing the medical codes to its descriptions.
  The text representations are encoded independently by the event encoder, and aggregated by the event aggregator to make a prediction.
  Since UniHPF treats EHRs as free text, this is capable of handling heterogeneous EHR systems with a single model.
  Note that the sub-word tokenizer and word embedding layer is omitted in this figure.
  }}}
  {\makebox[0.5\textwidth][c]{\includegraphics[trim={185 230 160 200},clip,width=1.2\textwidth]{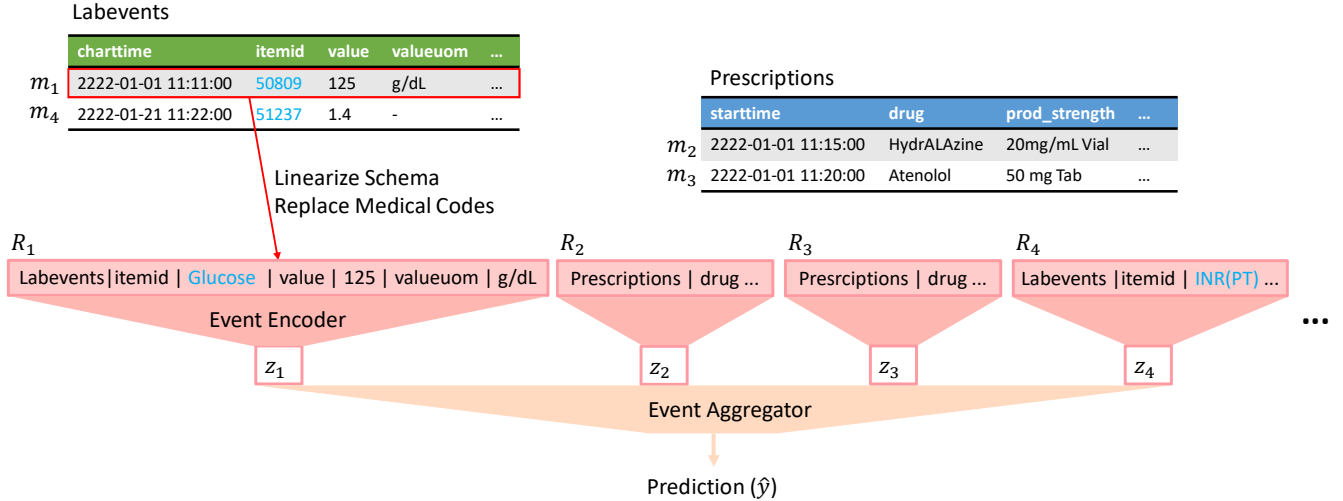}}}
  
\end{figure*}
\section{Background and Methods}
\label{sec:related}

\subsection{Federated Learning}
Federated learning is a kind of distributed learning that trains a model without sharing data among clients \citep{fedavg} (\algorithmref{alg:fed}).
It enables to train the model without the hazard of data leakage by aggregating the parameters or gradients \citep{brisimi2018federated}.
An obstacle to applying FL is that the data among clients is often not independent and identically distributed (non-i.i.d.).
This makes optimizing a global model challenging \citep{rieke2020future}.
We examine four well-known FL algorithms with UniHPF.

\camr{
\begin{itemize}
    \item \texttt{FedAvg} \citep{fedavg}, a \textit{de facto} algorithm of FL, simply averages the local model weights.
    Since this method does not fully consider the non-i.i.d. problem, various FL algorithms have been developed.
    \item \texttt{FedProx} \citep{fedprox} regularizes the local model with $L_2$ distance between local and global parameters.
    It prevents the weights of the local optimal points from taking the global point.
    \item \texttt{FedBN} \citep{fedbn} handle the feature heterogeneity among clients, by excluding the batch normalization layers from the aggregation step.
    \item \texttt{FedPxN} \citep{fedpxn} combined the advantages of \texttt{FedProx} and \texttt{FedBN}, and it is reported to show best performance for FL with EHRs.
\end{itemize}
}

\subsection{UniHPF}
As mentioned earlier, the EHR system heterogeneity is the biggest obstacle to performing multi-source learning. 
To overcome this problem, unifying the input format is required.
Recently, UniHPF \citep{unihpf} has successfully addressed this problem without using domain knowledge and excessive preprocessing (\figureref{fig:unihpf}).

The two key concepts of UniHPF are treating EHR as free text, and utilizing the EHR hierarchy.
A patient $\mathcal{P}$ in any EHR system is composed of multiple medical events $m_i \in \mathcal{P}$, and each event has its type $e_i$, such as ``labevents'' or ``prescriptions''.
The events are composed of the corresponding features, which is composed of name and value $(n_{i,j}, v_{i,j}) \in m_i$.
Some of the values are in the form of the medical codes $c$, and these differ among the EHR systems.
Thus, UniHPF replace the code $c$ with its text description $d$ \citep{descemb}.
For example, the lab measurement code ``50912'' can be converted into ``Glucose''.
UniHPF makes a free text representation $R_i$ of each event $m_i$ by linearizing the schema as
\[R_i = (e_i \oplus n_{i,1} \oplus v_{i,1} \oplus n_{i,2} \oplus v_{i,2} \oplus \cdots)\],
where $\oplus$ is a concatenation operator.
Note that UniHPF does not perform the feature selection, which is time- and cost-consuming.
Since these text representations are mutually compatible among the heterogeneous EHR systems, UniHPF is a suitable framework to perform FL.
To make a prediction $\hat{y}$, UniHPF uses sub-word tokenizer ($\text{Tok}$) and word embedding layer ($\text{Emb}$), and encodes the text-represented events individually with the event encoder ($\text{Enc}$).
\[z_i = \text{Enc}(\text{Emb}(\text{Tok}(R_i)))\]
The encoded events are aggregated by the event aggregator ($\text{Agg}$).
\[\hat{y} = \text{Agg}(z_1, z_2,\cdots)\]
This helps the model to understand the patient-event level hierarchy of EHRs.
Since no medical domain knowledge is used in any of the above steps, UniHPF can unify heterogeneous EHR systems efficiently.

\begin{figure*}[htbp]
\floatconts
    {fig:result}
    {\caption{Test AUPRC of the local learning (LL), federated learning (FL), and centralized learning (CL) experiments. Note that the graphs are ordered by the client size from left top to right bottom. $\star$ mark indicates the p-value of the Student's t-test is lower than 0.05.}}
    {\makebox[\textwidth][c]{{\includegraphics[width=1.2\linewidth]{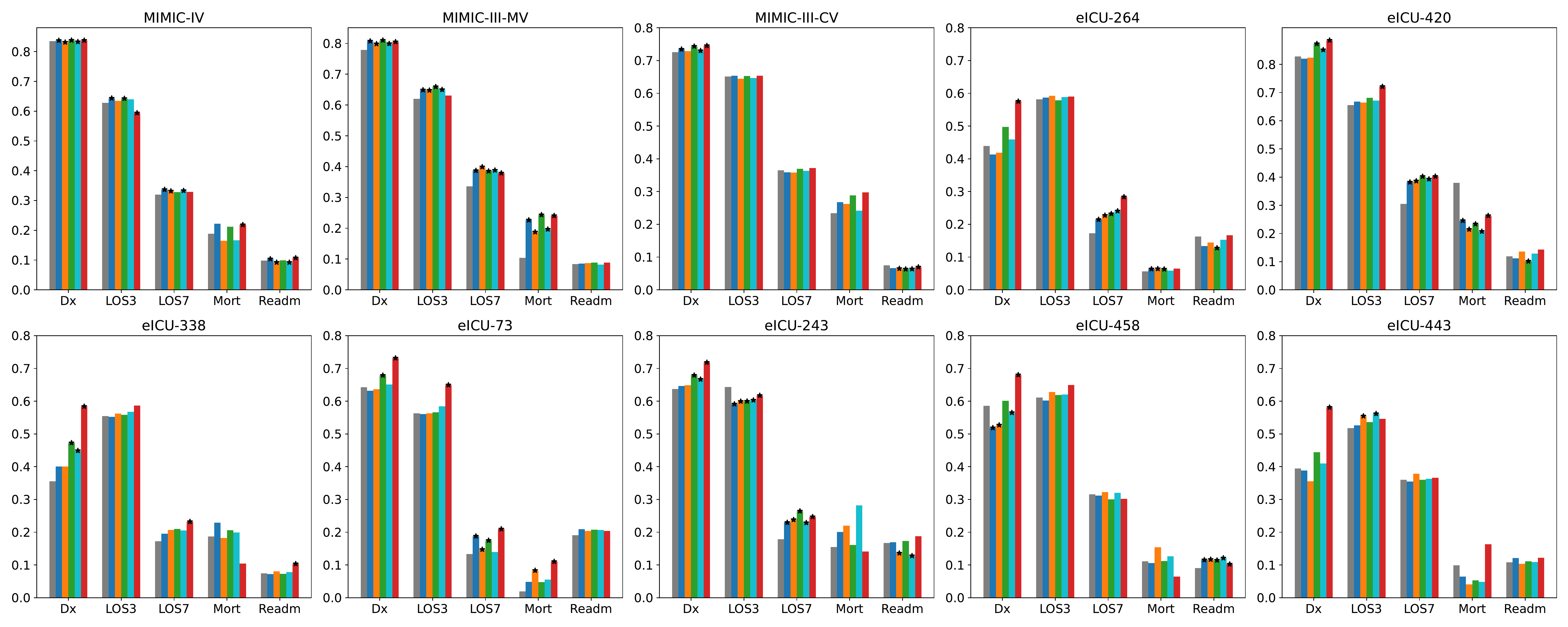}}}
    \includegraphics[width=1.0\linewidth]{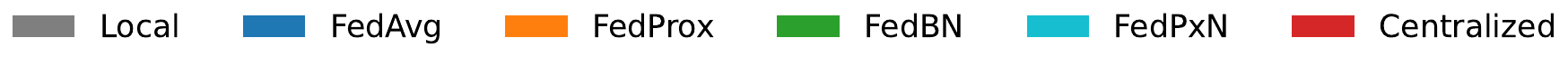}}
\end{figure*}

\section{Experiments and Discussion}

\subsection{Datasets}
We use three open-sourced EHR datasets: MIMIC-III \citep{mimic3}, MIMIC-IV \citep{mimic4}, and Philips eICU \citep{eicu}. The first two are composed of data from a single hospital, and the last one is a combination of data from multiple hospitals.
MIMIC-III is recorded with two heterogeneous EHR systems, so we split it into MIMIC-III-CV (CareVue) and MIMIC-III-MV (Metavision) based on the systems.
Since different hospitals have different data distributions, we treat the 7 largest hospitals in eICU dataset as independent clients.
To summarize, we have a total of 10 clients from 4 different EHR systems and 10 different cohorts
: MIMIC-IV, MIMIC-III-CV, MIMIC-III-MV, and 7 hospitals in eICU. 
Note that the clients are heterogeneous enough in terms of the demographic information and label distributions (\appendixref{apd:stat}).
The data is split into train, valid, and test set with 8:1:1 ratio in a stratified manner for each task.

\subsection{Experimental Setting}
Our cohorts include the patients over 18 years \ecedit{of age} who stayed in intensive care unit (ICU) longer than 24 hours. 
We only use the first 12 hours of the first ICU stay from each hospital admission to make predictions. We follow the settings of \citet{unihpf}, except that we use GRU \citep{chung2014empirical} as the event encoder of UniHPF.
All experimental resources \camr{and hyperparameters} are available on github\footnote{\url{https://github.com/starmpcc/UniFL}}.
We adopt 5 prediction tasks from \citet{mcdermott2020comprehensive}.
\begin{itemize}
    \setlength\itemsep{3pt}
    \item Diagnosis (Dx): Predict all categorized diagnosis codes during the whole hospital stay of a patient.
    \item Length of Stay (LOS3, LOS7): Predict whether a patient would stay in ICU longer than 3 or 7 days.
    \item Mortality (Mort): Predict whether a patient would be alive or die within 60 hours.
    \item Readmission (Readm): Predict whether a patient would readmit to ICU within the same hospital admission.
\end{itemize}

\camr{To compare with multi-source learning, we examine the performance of Local Learning (LL), which is training and evaluating with each client's data alone.}
We used Area Under Precision-Recall Curve (AUPRC) as the metric.
We assume a stable internet connection and full participation because the hospitals are generally connected by LAN.
All the experiments are repeated with five random seeds with one NVIDIA A100 80G or two RTX A6000 48G gpus. 

\subsection{Experimental Result}
The experimental results are shown \ecedit{in} \figureref{fig:result}. 
The average performance for each task and client are reported in \appendixref{apd:res}.
First, we evaluate whether our framework successfully handles the EHR system heterogeneity.
Second, we compare the FL algorithms with respect to the non-i.i.d. problem.

Consistent with \citet{unihpf}, UniHPF always gets an average of 10.4\% performance increase by using CL compared with LL.
This means that UniHPF properly overcomes the EHR system heterogeneity.

Overall, the FL shows an average of 3.4\% performance increase compared to LL.
This implies that UniHPF is helpful in aggregating the clients' data into a single FL model, demonstrating its potential since this combined method is practically applicable in real-world heterogeneous EHRs. 

We compare the performance among the algorithms with respect to the non-i.i.d. problem.
Our results agree with \citet{choudhury2019predicting, niu2020billion}, which showed CL is the upper bound of the FL performance.
In the CL setting, the clients' data is pooled before start the training, which prevents the non-i.i.d. problem.
Therefore, CL has the best performance among the learning methods.
\camr{In contrast, \texttt{FedAvg} can be treated as an empirical lower bound among the FL algorithms with some non-i.i.d. data \citep{li2019convergence, hsu2019measuring}.}
Although \texttt{FedProx} is an algorithm that handles the non-i.i.d. problem, the performance is lower than \texttt{FedAvg}.
The reason for this result seems that the performance of \texttt{FedProx} heavily depends on the hyperparameter $\mu$.
The performance of \texttt{FedBN} and \texttt{FedPxN} are higher than \texttt{FedAvg}, and lower than CL.
This suggests that these algorithms do address the non-i.i.d. problem in EHRs to some extent, but not completely.

Nevertheless, our framework shows its potential when the data distribution is extremely heterogeneous.
\ecedit{Contrary to } the other clients, eICU-73 and eICU-443 do not have drug infusion information.
\camr{Even in these extreme cases, performing FL with \texttt{FedBN} or \texttt{FedPxN} resulted in some performance increment compared to LL.}

For the training time, FL requires an average of 1.9 times more communication rounds than CL epochs until satisfying the same early stopping criterion.
Nevertheless, the performance of FL is inferior to CL. 
This result denotes that the gradient update is relatively less accurate for each communication round.
We expect that this can be improved by developing a better EHR-specific FL algorithm.

\section{Conclusion}

In this paper, we empirically show that the combination of UniHPF and FL successfully resolves both the EHR system heterogeneity and the non-i.i.d. problem simultaneously. 
The lower performance of FL compared to CL implies that there is still a room for improvement with a new FL algorithm in EHR.
We leave the investigation of EHR-specific pretraining with FL as our future work.



\acks{This work was supported by Institute of Information \& Communications Technology Planning \& Evaluation (IITP) grant (No.2019-0-00075), Korea Medical Device Development Fund grant (Project Number: 1711138160, KMDF\_PR\_20200901\_0097), and the Korea Health Industry Development Institute (KHIDI) grant (No.HR21C0198), funded by the Korea government (MSIT, MOTIE, MOHW, MFDS).}
\clearpage
\bibliography{pmlr-sample}

\onecolumn

\appendix

\section{Federated and Centralized Learning Algorithms}\label{apd:first}
In federated learning (\algorithmref{alg:fed}), each client has an individual model weight copy. 
The copies are initialized with the same weights and trained locally with corresponding client's data.
After the local training, the weights of the clients are gathered into the central server, aggregated, and synchronized.
Each FL algorithms have different corresponding \texttt{Aggregate} function.

\begin{figure*}[hb]
\begin{minipage}[t]{0.5\textwidth}
\begin{algorithm2e}[H]
\caption{Centralized Learning}
\label{alg:cl}
\KwIn{clients $C_1,\cdots,C_N$, data $D_1,\cdots,D_N$, total epochs $T$, learning rate $\eta$}
\KwOut{model parameter $w_{T}$}
\SetKwProg{Main}{main}{:}{}
\Main{}{
    \For {$i=1$ \KwTo $N$}{
        Copy $D_i$ to server\;
    }
    initialize server model with $w$\;
    $\mathcal{D}$ = \texttt{shuffle}$(D_1,\cdots, D_N)$\;
    
    \For {$t=0$ \KwTo $T-1$}
    {
        \For {\textup{batch} $b \leftarrow (x,y)$ \textup{of} $\mathcal{D}$}{
            $w \leftarrow w- \eta \nabla \mathcal{L}(w;b)$\;
        }
    }
    \Return $w$
}
\end{algorithm2e}

\end{minipage}
\begin{minipage}[t]{0.5\textwidth}
\begin{algorithm2e}[H]
\caption{Federated Learning}
\label{alg:fed}
\KwIn{clients $C_1,\cdots,C_N$, data $D_1,\cdots,D_N$, total communication rounds $T$, local epochs $L$, learning rate $\eta$}
\KwOut{model parameter $w_{T}$}
\SetKwProg{Main}{main}{:}{}
\SetKwProg{Def}{def}{:}{}
\SetKwFunction{FlocalT}{LocalTrain}
\Main{}{
    initialize client models with $w_0$\;
    \For {$t=0$ \KwTo $T-1$}
    {
        \For {$i=1$ \KwTo $N$}
        {
            $w_{t,i} \leftarrow w_t$\;
            $w_{t,i} \leftarrow$\texttt{LocalTrain} $(w_{t,k}, D_i)$\;
        }
        $w_{t+1}=$\texttt{Aggregate}$(w_1,\cdots,w_N)$\;
    }
    \Return $w_T$
}

\Def{\FlocalT{w,D}}{
    \For {$l=0$ \KwTo $L-1$}{
        \For{\textup{batch} $b \leftarrow (x,y)$ \textup{of} $D_k$}{
            $w \leftarrow w-\eta\nabla \mathcal{L}(w;b)$\;
        }
    }
    \Return $w$\;
}
\end{algorithm2e}
\end{minipage}
\end{figure*}

\clearpage

\section{Clients Statistics}\label{apd:stat}

\begin{table}[!htp]\centering
\caption{Cohort Statics and Label Distributions}\label{tab:stat}
\scriptsize
\setlength{\tabcolsep}{2pt}
\adjustbox{max width=\textwidth}{
\begin{tabular}{lrrrrrrrrrrrrrr}\toprule
\multicolumn{2}{c}{} &MIMIC-IV &MIMIC-III-MV &MIMIC-III-CV &eICU-264 &eICU-420 &eICU-338 &eICU-73 &eICU-243 &eICU-458 &eICU-443 &Micro Avg. &Macro Avg. \\\cmidrule{1-14}
\multicolumn{2}{c}{Cohort Size} &65594 &21160 &16831 &3637 &3153 &2636 &2612 &2423 &2368 &2367 &\multicolumn{2}{c}{12278.10} \\\cmidrule{1-14}
\multicolumn{2}{c}{No. of Unique codes} &1908 &1923 &3226 &347 &320 &367 &384 &306 &284 &281 &1844.43 &934.60 \\\cmidrule{1-14}
\multicolumn{2}{c}{Average No. of events per sample} &112.89 &91.43 &100.46 &53.69 &87.60 &47.32 &58.13 &51.80 &58.09 &51.82 &99.07 &71.32 \\\cmidrule{1-14}
\multicolumn{2}{c}{\textbf{Demographic Informations}} & & & & & & & & & & & & \\\cmidrule{1-2}
\multicolumn{2}{c}{Mean Ages} &63.28 &75.18 &73.90 &62.92 &63.73 &61.84 &63.54 &63.77 &61.60 &55.06 &66.58 &64.48 \\\cmidrule{1-14}
\multirow{2}{*}{Gender(\%)} &M &55.81 &56.31 &56.74 &51.61 &57.78 &55.73 &54.98 &55.57 &54.18 &57.02 &55.92 &55.57 \\\cmidrule{2-14}
&F &43.69 &43.69 &43.26 &48.39 &42.22 &44.27 &45.02 &44.43 &45.82 &42.98 &44.08 &44.38 \\\cmidrule{1-14}
\multirow{5}{*}{Ethnicity(\%)} &White &67.78 &72.78 &70.67 &87.82 &86.08 &92.87 &75.61 &64.05 &64.02 &42.97 &70.18 &72.46 \\\cmidrule{2-14}
&Black &10.84 &10.25 &8.74 &7.26 &4.19 &1.52 &13.51 &31.82 &29.10 &52.68 &11.61 &16.99 \\\cmidrule{2-14}
&Hispanic &3.82 &4.07 &2.80 &0.33 &0.03 &1.29 &7.66 &0.00 &0.00 &1.10 &3.35 &2.11 \\\cmidrule{2-14}
&Asian &2.96 &2.72 &2.10 &0.82 &1.49 &0.27 &1.30 &0.99 &1.27 &0.38 &0.00 &1.43 \\\cmidrule{2-14}
&Other &14.60 &10.18 &15.69 &3.77 &8.21 &4.06 &1.91 &3.14 &5.62 &2.87 &14.87 &7.00 \\\cmidrule{1-14}
\multicolumn{2}{c}{\textbf{Label Ratio}} & & & & & & & & & & & & \\\cmidrule{1-2}
\multirow{18}{*}{Dx(\%)} &1 &4.73 &4.81 &4.99 &3.78 &3.26 &0.48 &4.49 &3.40 &3.10 &3.85 &3.00 &3.69 \\\cmidrule{2-14}
&2 &3.99 &4.25 &4.16 &2.56 &1.75 &1.54 &2.59 &1.98 &1.23 &4.84 &1.30 &2.89 \\\cmidrule{2-14}
&3 &10.36 &10.87 &12.16 &5.55 &12.40 &8.87 &12.25 &11.75 &5.04 &7.09 &3.43 &9.63 \\\cmidrule{2-14}
&4 &6.77 &6.55 &6.24 &2.47 &8.97 &1.67 &3.33 &3.23 &1.55 &1.25 &2.38 &4.20 \\\cmidrule{2-14}
&5 &7.73 &6.60 &5.16 &2.49 &5.35 &1.90 &2.67 &2.11 &1.98 &2.55 &2.81 &3.85 \\\cmidrule{2-14}
&6 &6.18 &5.76 &4.27 &9.21 &5.83 &6.12 &4.82 &5.98 &6.63 &8.69 &2.34 &6.35 \\\cmidrule{2-14}
&7 &11.15 &11.92 &15.35 &23.57 &11.55 &22.44 &18.75 &24.62 &23.34 &19.63 &4.38 &18.23 \\\cmidrule{2-14}
&8 &6.95 &7.52 &9.24 &17.27 &9.91 &18.52 &12.81 &13.47 &15.40 &15.92 &2.96 &12.70 \\\cmidrule{2-14}
&9 &7.25 &7.43 &7.46 &6.94 &5.95 &5.99 &4.19 &3.88 &4.96 &3.92 &3.26 &5.80 \\\cmidrule{2-14}
&10 &6.98 &7.33 &7.53 &5.29 &6.93 &6.72 &9.58 &7.56 &12.60 &4.51 &3.30 &7.50 \\\cmidrule{2-14}
&11 &0.08 &0.05 &0.07 &0.06 &0.03 &0.05 &0.09 &0.04 &0.12 &0.07 &0.04 &0.06 \\\cmidrule{2-14}
&12 &1.53 &1.72 &1.88 &0.55 &0.88 &1.14 &0.46 &0.42 &0.56 &0.24 &0.77 &0.94 \\\cmidrule{2-14}
&13 &4.34 &4.33 &2.96 &0.63 &0.58 &0.51 &0.46 &0.36 &0.40 &0.54 &2.20 &1.51 \\\cmidrule{2-14}
&14 &0.58 &0.56 &0.57 &0.00 &0.02 &0.00 &0.06 &0.03 &0.00 &0.14 &0.31 &0.20 \\\cmidrule{2-14}
&15 &0.01 &0.00 &0.00 &0.00 &0.01 &0.00 &0.00 &0.00 &0.00 &0.00 &0.01 &0.00 \\\cmidrule{2-14}
&16 &5.78 &6.52 &8.12 &14.89 &10.85 &20.57 &18.70 &14.50 &18.32 &23.59 &3.03 &14.18 \\\cmidrule{2-14}
&17 &6.67 &5.55 &3.81 &3.58 &5.58 &2.53 &2.20 &4.98 &3.83 &2.55 &3.71 &4.13 \\\cmidrule{2-14}
&18 &8.91 &8.25 &6.04 &1.40 &10.25 &1.36 &2.73 &1.93 &1.25 &1.02 &5.23 &4.31 \\\cmidrule{1-14}
LOS3(\%) &true &32.50 &36.93 &42.24 &42.86 &48.24 &38.66 &37.10 &39.50 &39.74 &45.59 &36.26 &40.34 \\\cmidrule{1-14}
LOS7(\%) &true &11.08 &12.51 &15.37 &12.48 &16.97 &11.87 &11.06 &10.94 &15.58 &17.66 &12.44 &13.55 \\\cmidrule{1-14}
Mort(\%) &true &1.68 &2.70 &2.84 &2.12 &3.55 &2.35 &0.54 &1.32 &3.04 &2.83 &2.11 &2.30 \\\cmidrule{1-14}
Readm(\%) &true &7.94 &5.81 &5.72 &8.83 &11.48 &8.35 &17.99 &12.30 &8.49 &10.31 &7.75 &9.72 \\
\bottomrule
\end{tabular}
}
\end{table}

\clearpage

\section{Experimental Result}\label{apd:res}
\begin{table*}[!htp]\centering
\caption{Average performance for each task and client. The numbers in the parentheses mean the relative performance improvement compared to the local learning (LL). Red and blue texts mean the negative and more than 10\% increments, respectively.}\label{tab:res}
\scriptsize
\renewcommand{\arraystretch}{1.5}
\adjustbox{max width=\textwidth}{
\begin{tabular}{lrrrrrrr}\toprule
&Local &FedAvg &FedProx &FedBN &FedPxN &Centralized \\\hline
Dx &0.622 &0.619 \textcolor{red}{($-$0.38\%)} &0.616 \textcolor{red}{($-$0.87\%)} &0.663 (+6.67\%) &0.641 (+3.08\%) &0.714 \textcolor{blue}{(+14.81\%)} \\\hline
LOS3 &0.603 &0.603 (+0.12\%) &0.609 (+1.07\%) &0.609 (+1.12\%) &0.613 (+1.81\%) &0.623 (+3.54\%) \\\hline
LOS7 &0.266 &0.295 \textcolor{blue}{(+11.33\%)} &0.299 \textcolor{blue}{(+12.67\%)} &0.302 \textcolor{blue}{(+13.88\%)} &0.297 \textcolor{blue}{(+11.94\%)} &0.312 \textcolor{blue}{(+17.51\%)} \\\hline
Mort &0.153 &0.167 (+9.33\%) &0.157 (+2.76\%) &0.162 (+5.76\%) &0.158 (+3.35\%) &0.166 (+8.90\%) \\\hline
Readm &0.117 &0.118 (+1.62\%) &0.116 \textcolor{red}{($-$0.34\%)} &0.116 \textcolor{red}{($-$0.75\%)} &0.116 \textcolor{red}{($-$0.59\%)} &0.129 \textcolor{blue}{(+10.70\%)} \\
\hline \hline
MIMIC-IV &0.414 &0.428 (+3.59\%) &0.410 \textcolor{red}{($-$0.72\%)} &0.424 (+2.45\%) &0.412 \textcolor{red}{($-$0.24\%)} &0.417 (+0.83\%) \\\hline
MIMIC-III-MV &0.384 &0.431 \textcolor{blue}{(+12.15\%)} &0.423 \textcolor{blue}{(+10.33\%)} &0.437 \textcolor{blue}{(+13.81\%)} &0.423 \textcolor{blue}{(+10.10\%)} &0.428 \textcolor{blue}{(+11.61\%)} \\\hline
MIMIC-III-CV &0.41 &0.416 (+1.52\%) &0.411 (+0.38\%) &0.423 (+3.27\%) &0.409 \textcolor{red}{($-$0.24\%)} &0.427 (+4.23\%) \\\hline
eICU-264 &0.283 &0.282 \textcolor{red}{($-$0.06\%)} &0.289 (+2.36\%) &0.299 (+6.12\%) &0.300 (+6.17\%) &0.336 \textcolor{blue}{(+18.95\%)} \\\hline
eICU-420 &0.457 &0.445 \textcolor{red}{($-$2.58\%)} &0.445 \textcolor{red}{($-$2.67\%)} &0.458 (+0.25\%) &0.450 \textcolor{red}{($-$1.49\%)} &0.483 (+5.65\%) \\\hline
eICU-338 &0.269 &0.289 (+7.89\%) &0.286 (+6.60\%) &0.304 \textcolor{blue}{(+13.14\%)} &0.300 \textcolor{blue}{(+11.67\%)} &0.322 \textcolor{blue}{(+19.86\%)} \\\hline
eICU-73 &0.31 &0.327 (+5.77\%) &0.326 (+5.46\%) &0.334 (+8.13\%) &0.327 (+5.72\%) &0.380 \textcolor{blue}{(+23.00\%)} \\\hline
eICU-243 &0.356 &0.367 (+3.15\%) &0.368 (+3.43\%) &0.375 (+5.39\%) &0.381 (+7.14\%) &0.382 (+7.29\%) \\\hline
eICU-458 &0.343 &0.330 \textcolor{red}{($-$3.59\%)} &0.349 (+1.96\%) &0.349 (+1.84\%) &0.350 (+2.26\%) &0.359 (+4.92\%) \\\hline
eICU-443 &0.296 &0.291 \textcolor{red}{($-$1.52\%)} &0.286 \textcolor{red}{($-$3.12\%)} &0.300 (+1.76\%) &0.298 (+0.96\%) &0.355 \textcolor{blue}{(+20.33\%)} \\
\hline \hline
Average &0.352 &0.361 (+2.54\%) &0.359 (+2.19\%) &0.370 (+5.29\%) &0.365 (+3.76\%) &0.389 \textcolor{blue}{(+10.57\%)} \\
\bottomrule
\end{tabular}
}
\end{table*}

\end{document}